# A NOVEL APPROACH FOR GENERATING FACE TEMPLATE USING BDA

Shraddha S. Shinde[1] and Prof. Anagha P. Khedkar[2]

[1]P.G. Student, Department of Computer Engineering, MCERC, Nashik (M.S.), India.
shraddhashinde@gmail.com
[2]Associate Professor, Department of Computer Engineering, MCERC, Nashik (M.S.), India
anagha_p2@yahoo.com

## ABSTRACT

*In identity management system, commonly used biometric recognition system needs attention towards issue of biometric template protection as far as more reliable solution is concerned. In view of this biometric template protection algorithm should satisfy security, discriminability and cancelability. As no single template protection method is capable of satisfying the basic requirements, a novel technique for face template generation and protection is proposed. The novel approach is proposed to provide security and accuracy in new user enrollment as well as authentication process. This novel technique takes advantage of both the hybrid approach and the binary discriminant analysis algorithm. This algorithm is designed on the basis of random projection, binary discriminant analysis and fuzzy commitment scheme. Three publicly available benchmark face databases are used for evaluation. The proposed novel technique enhances the discriminability and recognition accuracy by 80% in terms of matching score of the face images and provides high security.*

## 1. INTRODUCTION

Biometric systems are being deployed in various applications including travel and transportation, financial institutions, health care, law enforcement agencies and border crossing, thus enhancing security and discriminability of biometric template. The human face is a feature that can be used by biometric systems. Human face recognition by analyzing the size and position of different facial features is being pushed for use at several airports to increase security. In spite of many advantages, biometric systems like any other security applications are vulnerable to a wide range of attacks. An attack on a biometric system can take place for three main reasons:

A person may wish to disguise his own identity. For instance, an individual/terrorist attempting to enter a country without legal permission may try to modify his biometric trait or conceal it by placing an artificial biometric trait (e.g. a synthetic fingerprint, mask, or contact lens) over his biometric trait. Recently, in January 2009, the Japanese border control fingerprint system was deceived by a woman who used tape-made artificial fingerprints on her true fingerprints.

An attack on a biometric system can occur because an individual wants to attain privileges that another person has. The impostor, in this case, may forge biometric trait of genuine user in order to gain the unauthorized access to systems such as person's bank account or to gain physical access to a restricted region.

A benefit to sharing biometric trait may be the cause to attack the biometric systems. Someone, for instance, can establish a new identity during enrollment using a synthetically generated biometric trait. Thus, sharing the artificial biometric trait leads to sharing that fraudulent identity with multiple people. To enhance the recognition performance as well as to provide the better security, new system is to be proposed. The proposed system is designed

- To enhance the discriminability of face template by using Binary discriminant analysis.
- To provide the better security to binary template against smart attacks and brute force attack.

## 2. EXISTING TEMPLATE PROTECTION SCHEME

Template Protection Scheme can be categorized into three main approaches: 1) the biometric cryptosystem approach 2) the transform-based approach and 3) hybrid approach. Figure 1 shows the categorization of template protection scheme. The basic idea of these approaches is that instead of storing the original template, the transformed/encrypted template which is intended to be more secure, is stored. In case the transformed/encrypted template is stolen or lost, it is computationally hard to reconstruct the original template and to determine the original raw biometric data simply from the transformed/encrypted template.

In the biometric cryptosystem approach, the error-correcting coding techniques are employed to handle intra-class variations. Two popular techniques, namely fuzzy commitment scheme [7] and fuzzy vault scheme [8], are discussed. The advantage of this approach is that, since the output is an encrypted template, its security level is high. However, the error-correcting ability of these schemes may not be strong enough to handle large intra-class variations such as face images captured under different illuminations and poses. Also, this approach is not designed to be revocable. Finally, the error-correcting coding techniques require input in certain format (e.g., binary strings or integer vectors with limited range), and it is hard to represent every biometric template in this desired format. In the transform-based approach, a transformed template is generated using a "one-way" transform and the matching is performed in the transformed domain. The transform-based approach has a good cancelability (revocability) property, but the drawback of this approach is the trade-off between performance and security of the transformed template. The hybrid approach retains the advantages of both the transform-based approach and biometric cryptosystem approach, and overcomes the limitations of individual approaches [9] [10].

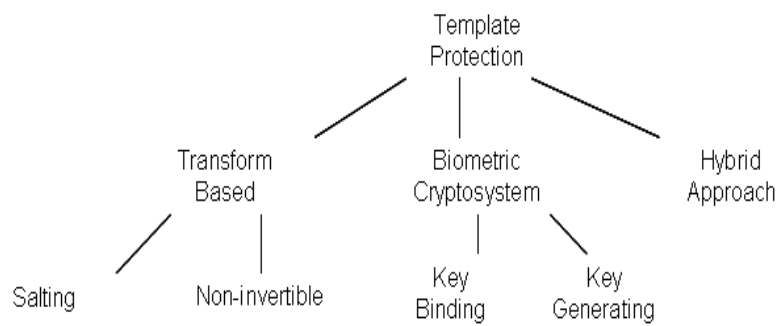

Figure 1. Categorization of template protection scheme

Some of the existing biometric template protection schemes provide security to binary template and rest of the other enhances the discriminability of the template. The available biometric cryptosystem and transformed based template protection schemes are not yet sufficiently mature for large scale deployment; they do not meet the requirements of diversity, revocability, security and high recognition performance. So in order to take the benefits of both approaches while eliminating their limitations, a hybrid approach for face biometric was developed only for the verification process but not for the new user enrollment process.

This limitation imposed to develop a new system to generate secure and discriminant face template for new user enrollment process as well as for the verification system.

## 3. ARCHITECTURE OF PROPOSED SYSTEM

A Novel Technique to generate discriminant and secure binary face template by using binary discriminating analysis is proposed. In this novel technique DP transform section from hybrid approach will replace by binary discriminant analysis. It is expected that this will enhance the discriminability of binary template. Finally, encryption algorithm is applied to generate secure template using fuzzy commitment. Figure 2 shows architecture flow of proposed system. Two stages (enrollment and authentication) of proposed system follow some steps which are explained in following section.

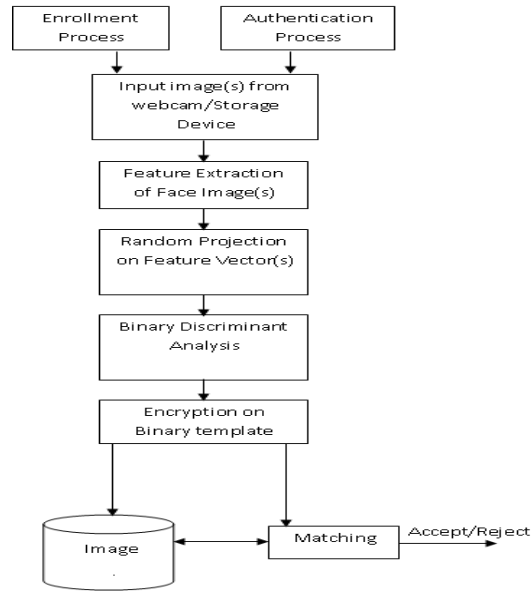

Figure 2. Architecture flow of proposed system

### 3.1. Random projection

Random projection is a popular dimensionality reduction technique and has been successfully applied in many computer vision and pattern recognition applications. Recently, it has also been employed as a cancelable transform for face biometric [9] [10]. The main purpose of the original random projection is to project a set of vectors into a lower dimensional subspace while preserving the Euclidean distances between vectors before and after the transformation, with a certain probability. The idea was first developed by Johnson and Lindenstrauss.

It will preserve the structure of the data without introducing significant distortion. It will represent faces in a random, low dimensional subspace. It uses random projection matrices to project data into low dimensional spaces. The matrix will be calculated using the following steps.

Step 1: Set each entry of the matrix to an independent and identically distributed (i.i.d.) value.
Step 2: Apply Gram-Schmidt algorithm to orthogonalize the rows of the matrix.

Step 3: To preserve the similarities in the low dimensional space normalize the rows of the matrix to unit length.

### 3.2. Binary discriminant analysis

BDA algorithm requires minimized within class variance and maximized between class variance. Perceptron method is used to find the optimal linear discriminant function. In the training phase, genuine label for each class is required by the perceptron method. The perceptron minimizes the distance between binary templates to the corresponding target binary template. This binary template is used as the reference for each class. This method is only used for minimization of within-class variance not for the between-class variance. To maximize the discriminability of the template between-class variance should be maximized.

But direct maximizing is difficult in binary space and conflicts with minimizing process of within-class variance. If the transformed binary templates in the same class are close to each other and binary templates of different class are far away then by using this new method it is possible to maximize discriminability of the binary template. The perceptron is used to find the optimal LDFs so that the output binary templates are closed to corresponding target binary template. Therefore within-class variance of binary template is minimized. Now, to maximize the between-class variance of the transform binary templates, target binary templates are randomly choose from the codeword of BCH codes having length, dimension with minimum distance. BCH codeword is randomly selected as the reference binary templates.

### 3.3. Fuzzy commitment

Fuzzy commitment scheme is applied on binary face template to provide the better security. Encryption algorithm is applied to encrypt the binary template and store binary template into database or to match with stored template.

## 4. RESULTS

This section discusses the results related discriminability and accuracy, computation time and security analysis for the hybrid approach and novel approach.

### 4.1. Discriminability and accuracy

Hybrid approach and novel approach generate there different face templates, namely cancelable template, binary template and secure template. In this section we evaluated the discriminability of each template in terms of the genuine and imposter histograms.

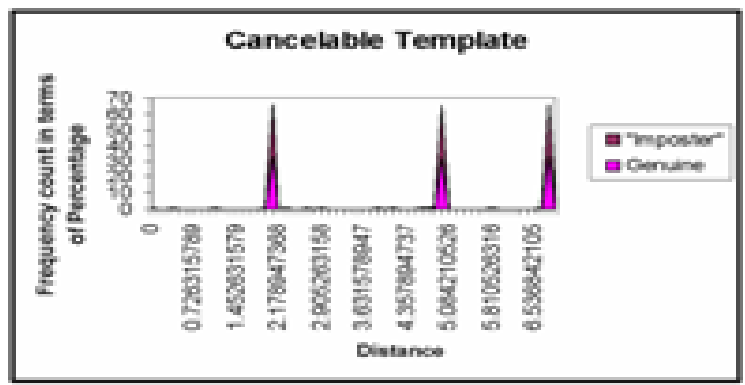

Figure 3. Histogram of cancelable template

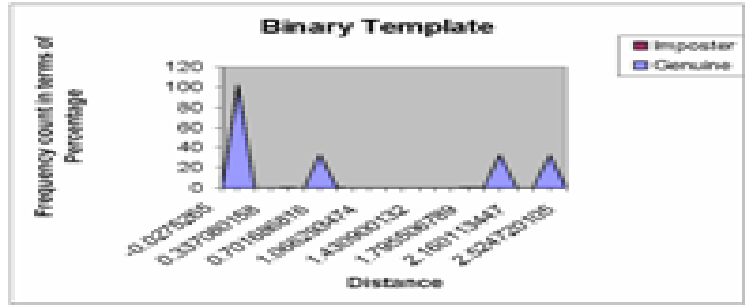

Figure 4. Histogram of binary template

In Particular, we illustrate that the cancelable template discriminability is enhanced in the binary template and matching score of the templates at each stage of hybrid and novel approach. Table number 1 and 2 shows the matching score of templates at each stage in hybrid as well as novel approach respectively. Figure 3 and 4 show histogram of cancelable template and binary template for imposter and genuine user. Matching score of feature vector is greater than cancelable template score that means the accuracy and discriminability is degraded in random projection stage. The degraded accuracy and discriminability is enhanced by the binary template generated using DP transform. Figure 5 shows the graphical representation of this result.

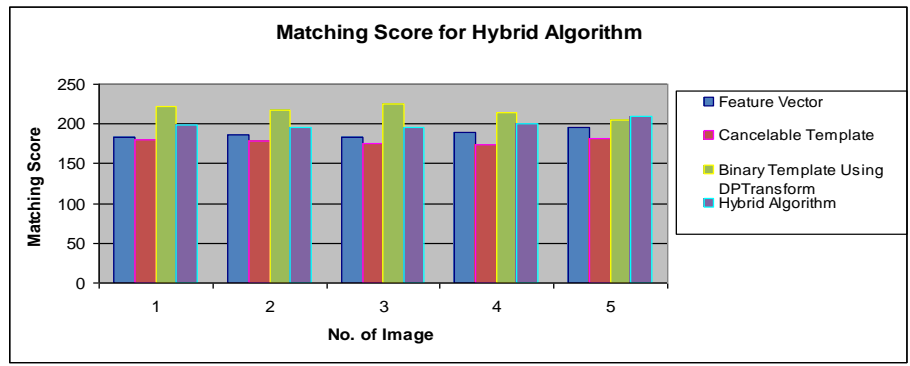

Figure 5. Matching score for hybrid approach

Matching score of feature vector (from Table 2) is greater than cancelable template score that means the accuracy and discriminability is degraded in random projection stage. The degraded accuracy and discriminability is enhanced by the binary template generated using binary discriminant analysis. Figure 6 shows the graphical representation of this result.

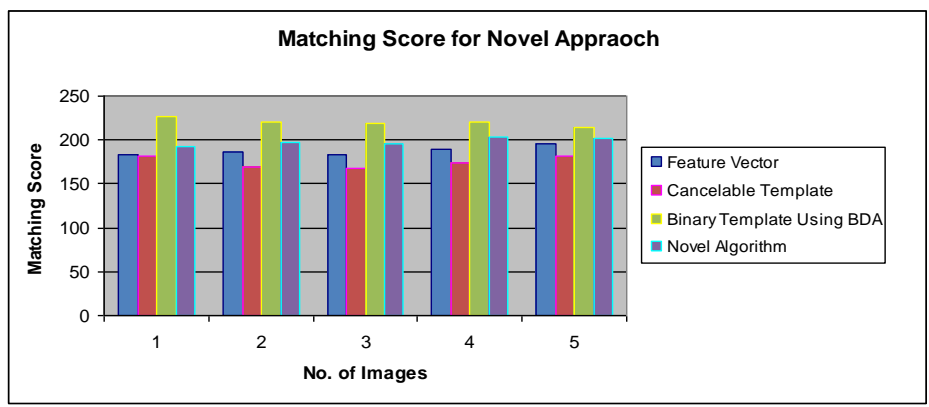

Figure 6. Matching Score for Novel Approach

Table 1 Matching score of five images for hybrid approach

| Images | Feature Vector | Cancelable Template | Binary Template Using DP Transform | Hybrid Algorithm |
|---|---|---|---|---|
| Image1 | 184 | 180 | 222 | 198 |
| Image2 | 187 | 178 | 218 | 196 |
| Image3 | 183 | 175 | 225 | 195 |
| Image4 | 189 | 174 | 215 | 201 |
| Image5 | 195 | 182 | 205 | 210 |

Table 2 Matching score of five images for a novel approach

| Images | Feature Vector | Cancelable Template | Binary Template Using BDA | Novel Algorithm |
|---|---|---|---|---|
| Image1 | 184 | 182 | 226 | 193 |
| Image2 | 187 | 170 | 220 | 197 |
| Image3 | 183 | 168 | 219 | 195 |
| Image4 | 189 | 174 | 221 | 203 |
| Image5 | 195 | 182 | 214 | 202 |

Table 3 Matching score for five images for DP transform and binary discriminant analysis

| Images | Binary Template Using DP Transform | Binary Template Using BDA |
|---|---|---|
| Image1 | 222 | 226 |
| Image2 | 218 | 220 |
| Image3 | 225 | 219 |
| Image4 | 215 | 221 |
| Image5 | 205 | 214 |

The table no. 3 shows the matching score for five images for DP transform and binary discriminant analysis. The comparison of both the stages shows matching score for templates generated by binary discriminant analysis is greater than DP transform. This matching score is up to 80%. Figure 7 shows the graphical representation of this result.

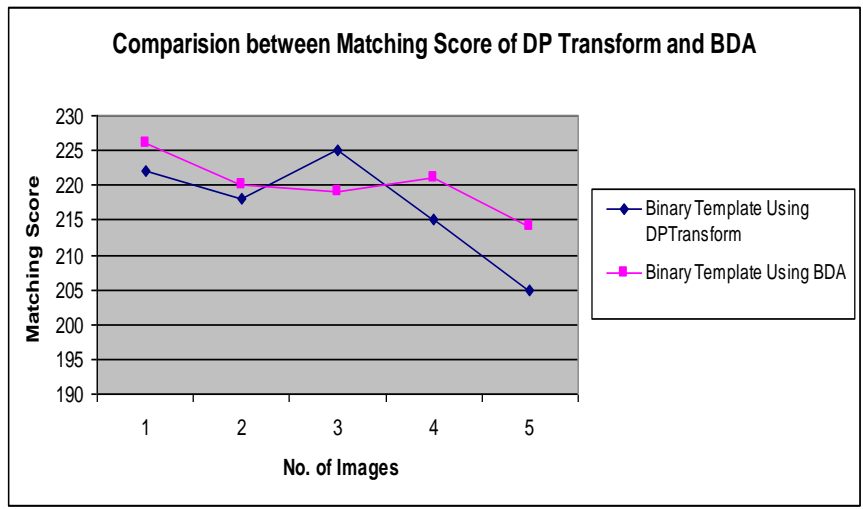

Figure 7. Comparison of matching score for novel approach and hybrid approach

## 4.2. Computation time

All the experiments are performed on typical personal computer having configuration as core i5 processor and both algorithms (hybrid and novel approach) are implemented using NetBeans IDE. Table no 4 and 5 shows the computation time for one class training process and the time for one input query template respectively. This time includes random projection, DP Transform and fuzzy commitment for hybrid approach as well for novel approach that includes random projection, BDA process and fuzzy commitment. Here we considered 10 sample images from standard database FERET, FRGC database and other data sets. Analysis of computation time of enrollment and verification stage is shown in Figure 8 and 9 respectively.

Table 4 Computation time for verification process for ten images (time in seconds)

| Data Set | For Hybrid Approach | For Novel Approach |
|---|---|---|
| FERET | 348 | 335 |
| FRGC | 366 | 370 |
| CMUPIE | 342 | 340 |
| OTHER | 372.6 | 365 |

Table 5 Computation time for verification process for single image (time in seconds)

| Data Set | For Hybrid Approach | For Novel Approach |
|---|---|---|
| FERET | 44 | 48 |
| FRGC | 40 | 39 |
| CMUPIE | 41 | 43 |
| OTHER | 43 | 43 |

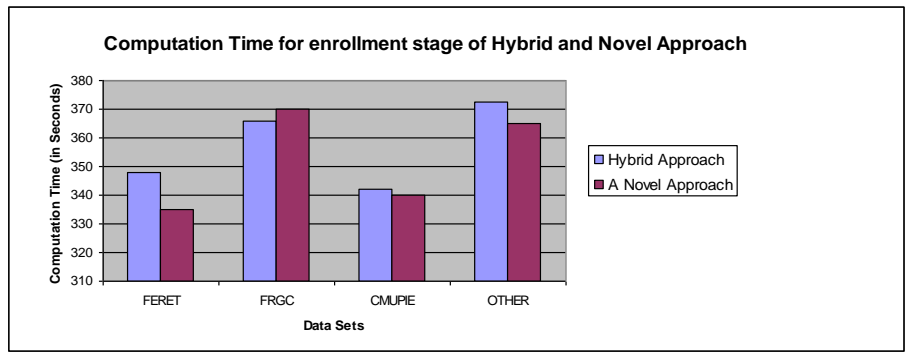

Figure 8. Analysis of computation time of enrollment stage

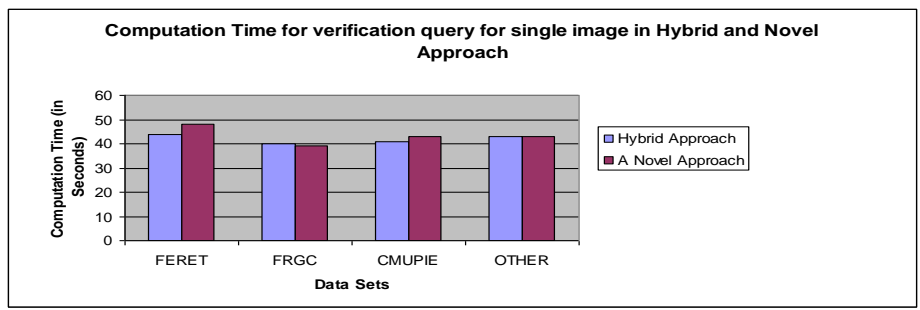

Figure 9. Analysis of computation time of verification stage

## 4.3. Security analysis

This section analyzes the security strength of the hybrid approach, a novel approach at each stage of these algorithms. Two types of potential attacks, namely brute force and "smart" attacks are considered. Brute force attack tries to guess the biometric data without any information, such as matching score, to attack the system. The smart attack viz. affine transformation attack is considered. It can be seen that the security strength of random projection and DP transform are low and medium respectively in smart attack, but full hybrid algorithm is secure against this attack.

In case of novel approach, it can be observed that the security strength of random projection and BDA process are low and medium respectively, but complete novel algorithm is highly secure. Hybrid approach and a novel approach are highly secure against brute force attack. In this case attacker requires number of trial of all possible combination of all alphanumerical character set. Here we have assumed character set of length 40 (including alphabets, numerical characters and other special characters). The attacker may try to guess the templates of each step.

### 4.3.1. Random projection

In hybrid algorithm, T1 is template generated at random projection step having length Kc (Kc is 3772). Therefore, it will cost the attacker $2^{kc-1}$ operations to guess it. So this step is secure against brute force attack.

### 4.3.2. DP transform

In this step, T2 is template generated at DP Transform having Kc distinguishing points and there are totally $2^{kc}$ combinations. Therefore, it will cost the attacker $2^{kc}$ combinations to guess it. With known distinguishing points it is hard to implement brute force attack against this step. So this step is secured against brute force attack. For the affine transformation attack, the real valued template is very hard to be reconstructed from a binary template. Moreover matching score, distinguishing points are not useful in this attack. Therefore, the DP transform is secured against an affine transformation attack.

### 4.3.3. Binary discriminant analysis

In this step, T2 is template generated at BDA having Kc length and there are totally $2^{kc-1}$ combinations. Therefore, it will cost the attacker $2^{kc-1}$ combinations to guess it. So this step is secure against brute force attack.
For the affine transformation attack, the real valued template is very had to be reconstructed from a binary template. Moreover matching score is not useful in this attack. Therefore, the BDA process is very secure against an affine transformation attack.

### 4.3.4. Fuzzy commitment scheme

B1 is template generated in this step having length Kc (Kc is 11340). Therefore, it will cost the attacker $2^{kc-1}$ operations to guess it. So this step is highly secured against brute force attack. Fuzzy Commitment Scheme performs matching operation between two hashed data. Because of property of the hash function, the distance between these two hash data will not reveal distance information. Therefore matching score is useless for affine transformation attack. Therefore, the Fuzzy Commitment Scheme is secured against an affine transformation attack.

### 4.3.5. Full algorithm

Since the three steps are integrated together to form hybrid algorithm, the attacker cannot get the output from these steps. Thus this algorithm does not reveal any information to attackers. In this algorithm, we have encrypted binary template which cannot be easily accessible to the attackers.

The output of each step is combined into one string in specific sequence is arranged. After this, the resultant string is converted into bytes which reduce the string length to store into database. Finally, this converted template is stored or matched with stored database template. Recovery of this stored template from brute force attack requires $2^{kc-1}$ (Kc is 6810) operations. Therefore, binary template recovery is not possible using affine transformation attack. So this full algorithm has high security strength against both attacks.

As the three steps are integrated together to form novel algorithm, the attacker cannot get the output from these steps. Thus this algorithm does not reveal any information to attackers. In this algorithm, we have encrypted binary template which cannot be easily accessible to the attackers. The output of each step is combined into one string in specific sequence is arranged. After this, this string is converted into bytes which reduce the string length to store into database. Finally, this converted template is stored or matched with stored database template. Recovery of this stored template from brute force attack requires $2^{kc-1}$ (Kc is 6800) operations.

Table 6 Security strength of the hybrid algorithm as well as each step in the full algorithm

| Attack | Random Projection | DP transform | Fuzzy Commitment | Full Algorithm |
|---|---|---|---|---|
| Brute Force | High | High | High | High |
| Affine Transformation | Low | High | High | High |

Table 7 Security strength of the novel algorithm as well as each step in the full algorithm

| Attack | Random Projection | BDA | Fuzzy Commitment | Full Algorithm |
|---|---|---|---|---|
| Brute Force | High | High | High | High |
| Affine Transformation | Low | High | High | High |

Therefore, binary template recovery is not possible using affine transformation attack. So this full algorithm has high security strength against both attacks. Table no. 6 and 7 show the security strength of hybrid algorithm and a novel approach at each stage.

## 4. CONCLUSIONS

A novel approach using BDA was designed, implemented and rigorously tested on the standard benchmark FERET, FRGC and CMU PIE database. Before developing this novel approach, hybrid approach was also tested on the similar database. The results of both the methods viz. novel approach and hybrid approach are compared in terms of discriminabilty and security. The proposed novel technique enhances the discriminability and recognition accuracy by 80% in terms of matching score of the face images and provides high security. This clearly indicates the performance improvement in novel approach against hybrid approach.

## REFERENCES

[1] N. Ratha, J. Connell, and R. Bolle, "Enhancing security and privacy in biometric based authentication systems", IBM Syst. J., vol. 40, no. 3, pp. 614-634, 2001.

[2] U. Uludag, S. Pankanti, S. Prabhakar, and A. K. Jain, "Biometric cryptosystems: Issues and challenges", Proc. IEEE, vol. 92, pp. 948-960, 2004.

[3]   L. O'Gorman, "Comparing passwords, tokens, and biometrics for user authentication", Proc. IEEE, vol. 91, pp. 2019-2040, 2003.

[4]   Adler, Images can be regenerated from quantized biometric match score data", Proc. Canadian Conf. Electrical and Computer Engineering, pp. 469-472, 2004.

[5]   Alder, Vulnerabilities in biometric encryption systems", Proc. IEEE Int. Conf. Audio- and Video-Based Biometric Person Authentication, vol. 3546, pp. 1100-1109, 2005.

[6]   K. Jain, K. Nandakumar, and A. Nagar, "Biometric template security", EURASIP J. Adv. Signal Process, 2008. [Online] Available:http://www.hindawi.com/journals/asp/2008/579416.abs.html

[7]   Juels and M. Wattenberg, "A fuzzy commitment scheme", Proc. Sixth ACM Conf. Comp. and Commun. Security, 1999, pp. 28-36.

[8]   Juels and M. Sudan, "A fuzzy vault scheme", IEEE Int. Symp. Information Theory, pp. 408, 2002.

[9]   Yi C. Feng, Pong C. Yuen, Member, IEEE, and Anil K. Jain, Fellow, IEEE, "A Hybrid Approach for Generating Secure and Discriminating Face Template", IEEE Transactions on Information Forensics and Security, Vol. 5, No. 1, March 2010.

[10]  Y. C. Feng, P. C. Yuen, and A. K. Jain, "A hybrid approach for face template protection", Proc. Int. Society for Optical Engineering (SPIE), 2008, vol. 6944, pp.1-11.

[11]  G. Davida, Y. Frankel, and B. Matt, "On enabling secure applications through online biometric identification", IEEE Symp. Privacy and Security, pp. 148-157, 1998.

[12]  F. Monrose, M. K. Reiter, and S. Wetzel, "Password hardening based on key stroke dynamics", Proc. ACM Conf. Computer and Communication Security, pp. 73-82,1999.

[13]  F. Monrose, M. Reiter, Q. Li, and S. Wetzel, "Cryptographic key generation from voice", Proc. IEEE Symp. Security and Privacy, pp. 202-213, 2001.

[14]  Goh and D. C. L. Ngo, "Computation of cryptographic keys from face biometrics", Proc. 7th IFIP TC6/TC11 Conf. Commun. Multimedia Security, vol. 22, pp. 1-13, 2003.

[15]  P. Tuyls and J. Goseling, "Capacity and examples of template-protecting biometric authentication systems", ECCV Workshop BioAW, pp. 158-170,2004.

[16]  T. Kevenaar, G. Schrijen, M. Veen, A. Akkermans, and F. Zuo, "Face recognition with renewable and privacy preserving binary templates", Proc. Fourth IEEE Automatic Identification Advanced Technologies, pp. 21-26, 2004.

[17]  F. Hao, R. Anderson, and J. Daugman, "Combining Cryptography with Biometric Effectively University of Cambridge", Tech. Rep. UCAM-CL-TR-640, ISSN 1476-2986, 2005.

[18]  Nagar, K. Nandakumar, and A. K. Jain, "Securing fingerprint template: Fuzzy vault with minutiae descriptors", Proc. Int. Conf. Pattern Recognition, pp. 1-4, 2008.

[19]  Y. Dodis, L. Reyzin, and A. Smith, "Fuzzy extractors: How to generate strong keys from biometrics and other noisy data", Proc. Advances in Cryptology-Eurocrypt, pp. 523-540, 2004.

[20]  D. Ngo, A. Teoh, and A. Goh, "Biometric hash: High-con_dence face recognition", IEEE Trans. Circuits Syst. Video Technol., vol. 16, no. 6, pp. 771-775, Jun. 2006.

[21]  E. C. Chang and S Roy, "Robust extraction of secret bits from minutiae", Proceedings of IEEE International Conference on Biometrics, pp. 750-759, 2007.

[22]  Abhishek Nagar, Karthik Nandakumar, Anil K. Jain, "A hybrid biometric cryptosystem for securing fingerprint minutiae templates", Pattern Recognition Letters 31, 733-741, 2010.

[23]     Y. C. Feng and P. C. Yuen, "Binary Discriminant Analysis for Face Template Protection", International Conference on Pattern Recognition, pp. 874-877, 2010.

[24]     Y. C. Feng and P. C. Yuen, "Binary Discriminant Analysis for Face Template Protection", IEEE Transaction, 2011.

[25]     N. Ratha, J. Connell, R. Bolle, and S. Chikkerur, "Cancelable biometrics: A case study in fingerprints", Proc. Int. Conf. Pattern Recognition, 2006, pp. 370-373.

[26]     N. Ratha, S. Chikkerur, J. Connell, and R. Bolle, "Generating cancelable fingerprint templates", IEEE Trans. Pattern Anal. Mach. Intell., vol. 29, no. 4, pp. 561-752, Apr. 2007.

[27]     S. Tulyakov, V. Chavan, and V. Govindaraju, "Symmetric hash functions for fingerprint minutiae", Proc. Int.Workshop Pattern Recognition for Crime Prevention, Security, and Surveillance, pp. 30-38, 2005.

[28]     R. Ang, R. Safavi-Naini, and L. McAven, "Cancelable key-based fingerprint templates", ACISP, pp. 242-252,2005.

[29]     Y. Sutcu, H. Sencar, and N. Nemon, "A secure biometric authentication scheme based on robust hashing", Proc. Seventh Workshop Multimedia and Security, pp.111-116, 2005.

[30]     Teoh, D. Ngo, and A. Goh, "Biohashing: Two factor authentication featuring fingerprint data and tokenised random number", Pattern Recognit., vol. 37, no. 11, pp. 2245-2255, 2004.

[31]     Teoh, A. Goh, and D. Ngo, "Random multispace quantization as an analytic mechanism for biohashing of biometric and random identity inputs", IEEE Trans. Pattern Anal. Mach. Intell., vol. 28, No. 12, pp. 1892-1901, Dec. 2006.

[32]     Wang Y, Plataniotis K, "Face based biometric authentication with changeable and privacy preservable templates", Proc of the IEEE Biometrics Symposium, 11-13, 2007.

[33]     Osama Ouda, Norimichi Tsumura and Toshiya Nakaguchi, "Tokenless Cancelable Biometrics Scheme for Protecting IrisCodes", International Conference on Pattern Recognition, 2010.

[34]     Y. C. Feng and P. C. Yuen, "Selection of distinguish points for class distribution preserving transform for biometric template protection", Proc. IEEE Int. Conf. Biometrics (ICB), pp. 636-645, 2007.

[35]     Y. C. Feng and P. C. Yuen, "Class-Distribution Preserving Transform for Face Biometric Data Security", Proceedings of IEEE International Conference on Acoustics, Speech, and Signal Processing (ICASSP), pp. 141-144, 2007.

[36]     Christian Rathgeb and Andreas Uhl, "Survey on biometric cryptosystems and cancelable biometrics", EURASIP Journal on Information Security, 2011.

[37]     P. N. Belhumeur, J. P. Hespanha, and D. J. Kriegman, Eigenfaces vs. fisher faces: Recognition using class specific linear projection", IEEE Trans. Pattern Anal. Mach. Intell., vol. 19, no. 7, pp. 711-720, Jul. 1997.